\documentclass[acmtog]{acmart}

\AtBeginDocument{%
  \providecommand\BibTeX{{%
    \normalfont B\kern-0.5em{\scshape i\kern-0.25em b}\kern-0.8em\TeX}}}

\setcopyright{acmcopyright}
\copyrightyear{2020}
\acmYear{2020}

\acmJournal{TKDD}

\usepackage{enumitem}
\graphicspath{ {./fig/} }
\begin{document}

\title{Continual Deterioration Prediction for Hospitalized COVID-19 Patients}


\author{Jiacheng Liu}
\authornote{University of Minnesota, Twin Cities. Minneapolis, Minnesota, USA}
\email{liu00520@umn.edu}
\orcid{0000-0002-1449-5038}

\author{Meghna Singh}
\authornotemark[1]

\author{Catherine St.Hill}
\authornote{Allina Health. Minneapolis, Minnesota, USA}
\author{Vino Raj}
\authornotemark[2]
\author{Lisa Kirkland}
\authornotemark[2]
\author{Jaideep Srivastava}
\authornotemark[1]

\renewcommand{\shortauthors}{}

\begin{abstract}
Leading up to August 2020, COVID-19 has spread to almost every country in the world, causing millions of infected and hundreds of thousands of deaths. In this paper, we first verify the assumption that clinical variables could have time-varying effects on COVID-19 outcomes. Then, we develop a temporal stratification approach to make daily predictions on patients’ outcome at the end of hospital stay. Training data is segmented by the remaining length of stay, which is a proxy for the patient's overall condition. Based on this, a sequence of predictive models are built, one for each time segment. Thanks to the publicly shared data, we were able to build and evaluate prototype models. Preliminary experiments show 0.98 AUROC, 0.91 F1 score and 0.97 AUPR on continuous deterioration prediction, encouraging further development of the model as well as validations on different datasets. We also verify the key assumption which motivates our method. Clinical variables could have time-varying effects on COVID-19 outcomes. That is to say, the feature importance of a variable in the predictive model varies at different disease stages.

\end{abstract}

\keywords{COVID-19, Machine Learning, Patients Triaging, Patients Deterioration Prediction}

\maketitle

\section{\textbf{Introduction}}
\subsection{\textbf{Background}}
The ongoing pandemic of COVID-19 has brought critical challenges to healthcare providers\cite{covid2020forecasting, garg2020hospitalization, wu2020risk}. Many COVID-19 positive individuals deteriorate rapidly after the initial onset of symptoms and require hospitalization and intensive care including the use of mechanical ventilators and other intensive supportive measures for survival. There is an urgent need for effective COVID-19-specific care protocols or decision-aiding tools addressing several key clinical problems, such as predicting clinical deterioration and risk of mortality, estimating length of stay, and predicting usage of mechanical ventilators\cite{bai2020predicting,haimovich2020development, jiang2020towards, white2020framework}. Ideally, given enough COVID-19 patients’ data, a data-driven approach would model these problems as either classification or regression tasks. However, this effort is largely impeded by the lack of privacy-preserved and law-compliant shared data. 
\subsection{\textbf{Related Work}}
Machine learning techniques have been applied to aid triaging decisions for COVID-19 patients and have shown promising results. Both \cite{liang2020development} and \cite{wu2020development} build logistic regression models to predict whether a patient will develop critical conditions such as admission to ICU, need of invasive ventilators or death. These models are based on clinical symptoms, lab results, radiology report, medical history and demographics. \cite{liang2020development} report an AUROC of 0.88 on independent test sets. \cite{wu2020development} validate their model on 5 cohorts collected across hospitals in Belgium, China and Italy. The AUROC ranges from 0.84 to 0.89. However, these models are one-shot classification models working at the time of admission, rather than models making daily predictions. \cite{yan2020interpretable} released a publicly available COVID-19 electronic medical records dataset consisting of 485 COVID-19 patients which can be used for various research purposes. We will refer to this dataset as \textit{Wuhan Dataset} in the rest of this paper, as all patient data is collected from Wuhan, China. In their work, XGBoost\cite{chen2016xgboost} models are built to predict risk of mortality for patients’ last day records. The result shows 97\% Accuracy on test and over 0.9 F1 score for both who survived and who died.  However, these retrospective models are not useful in an operational setting because they rely on lab results on the day of the final outcome. We argue that their retrospective study cannot be directly applied to real-world COVID-19 deterioration prediction scenarios, for two reasons, 
(i) It is impossible to know a patient’s exact outcome date in advance.
(ii) Variables may play time-varying roles during different stages of disease progression.
\subsection{\textbf{Contributions}}

Since ``retrospective models'' are not useful in an operational setting, we propose a temporally stratification approach to train ``prospective models'' that can make accurate predictions every day on the final outcome based on lab results, vitals, demographics and other medical data. The problem of predicting risk of mortality is known as the deterioration prediction problem. During the training phase of the proposed method, we stratify patients by their days until outcome, namely remaining length of stay. Secondly, a strata classifier is learned to predict the remaining length of stay for every patient. For each stratum, a risk of mortality model is built using all patient data that fall into this stratum. While testing and running, the predicted value is calculated as the dot product of strata probability distribution given by the strata classifier and the risk of mortality score given by each stratum model. 
Thus, our contributions are twofold.
\begin{enumerate}[label=(\roman*)]
    \item Verified the assumption that clinical variables play time-varying roles during different stages of disease progression.
    \item Proposed a temporal stratified model to solve the issue of time-varying effects.
\end{enumerate}

The rest of the paper is organized as follows. In section\ref{sec2}, we first examine the original problem definition used in \cite{yan2020interpretable}, then propose a new and more realistic definition of the deterioration prediction problem for COVID patients. Notations and abbreviations are also introduced there. In section\ref{sec3}, we present empirical evidence which support the claim that clinical variables have  time-varying effects. In section\ref{sec4}, we elaborate our two-step stratification-prediction method and experiment results are highlighted in section\ref{sec5}. Finally, drawbacks of this study and future work are discussed in section\ref{sec6}.

\section{\textbf{Problem Definition}}\label{sec2}
The main goal of this preliminary study is to predict whether patients will survive at end of their hospitalizations. This decision-making tool may ultimately assist first line hospital workers in prediction of outcomes for hospitalized COVID-19 patients, either admitted to general wards or the intensive care unit (ICU). Other predictive goals, such as estimated hospital length of stay, are also crucial for hospital resources management during the COVID-19 pandemic. 

\begin{table}
  \caption{Notations and Abbreviations}
  \label{tab:notation}
  \begin{tabular}{cl}
    \toprule
    Symbols & Meaning\\
    \midrule
    $LoS$ & Length of stay\\
    $DNR$  & Do not resuscitate\\
    $n_{p}$ & Number of patients\\
    $n_{s}$ & Number of strata or hidden states\\
    $X_{t,i}$ & Variable $X$ for patient $i$ on day $t$\\
    & Unless specified, $i$ is omitted.\\
    $X^{k}$ & Variable or function $X$ in stratum $k$\\
    $t \pm n$ & $n$ day(s) after or before day $t$ in hospital\\
    $LoS_{t}$ & Remaining length of stay on day $t$\\
    $T_{a}$ & Admission day\\
    $T_{e}$ & End of stay or discharged day\\
    $T_{LoS}$ & Total length of stay in days. \\
    & We have $T_{LoS} = T_{e} - T_{a} +1$\\
    $n_{d}$ & Number of days with new and non-NA clinical data.\\
    & $0 \leq n_{d} \leq T_{LoS}$\\
    $d_{i}$ & index the days when there are new and valid \\
    & clinical data.\\
    $\mathcal{L}$ & Loss function\\
    $\mathcal{S}$ & Temporal strata definition, which is essentially a set \\
    & of fixed time points segmenting the entire \\
    & hospital stay into different periods.\\
    $\mathcal{F}$ & Strata classifier\\
    $M^{k}$ & Stratum-level deterioration prediction model \\
    & for Stratum $k$\\
    $\mathbf{ x_{t}}$ & Feature vector on day $t$\\
    $p_{t}^{k}$ & Probability of being in stratum $k$ on day $t$\\
    $\hat{s}_{t}$ & Predicted stratum on day $t$\\
    $s_{t}$ & Actual stratum label on day $t$\\
    $\hat{y}_{t}^{k}$ or $M_{t}^{k}(\cdot)$ & End of stay outcome predicted by stratum model $k$ \\
    & on day $t$\\
    $\hat{y}_{t}$ & Predicted end of stay outcome on day $t$\\
    $y$ & The binary ground-truth outcome\\
  \bottomrule
\end{tabular}
\end{table}

We begin by introducing the notations, as shown in table\ref{tab:notation}. As line 5 in Table\ref{tab:notation} illustrates, the first subscription of a variable denotes time, namely days in the settings of this study\footnote{In theory, the choice of prediction frequency or time granularity can be arbitrary. It can be hourly, daily or several times a day or whenever there is new patient data available. We choose to do daily predictions here in alignment with the original work.}, while the second one denotes $i-th$ patient. It is worth noting that for simplicity, we omit the subscription of patient index most of the time, as all the operations can be trivially applied to every patient, except when discussing evaluation metrics of model performance, where subscriptions of patient index are explicitly needed. We reiterated the problem definition in \cite{yan2020interpretable} using the notations in Table\ref{tab:notation}.

\textbf{Problem Statement 0}\label{stmt:zero} (Retrospective Modeling of COVID-19 Deterioration)\footnote{Though not explicitly stated in the paper\cite{yan2020interpretable}, this can be inferred from the data preprocessing and model training/evaluation codes released in the github repo: \url{https://github.com/HAIRLAB/Pre_Surv_COVID_19}}:Given only COVID-19 patients’ laboratory results as input data, we build regularized binary classifier $C$ to minimize the cross entropy between actual outcome $y_{i}$ and the predicted outcome $\hat{y}_{t(i),i}=C(\mathbf{ x_{t(i),i}})$, where $t(i)=n_{d,i}$.

In the definition above, $x_{t(i),i}$ refers to the last available clinical data readings, or the closest data record to the discharged date, regardless of the method used to impute and process the data. There are three major limitations of this problem formulation. 
\begin{itemize}
    \item The entire dataset is underutilized. There are about 6,000 records in the raw dataset, however a majority of those are not used during model training since each patient will only have one record, \textit{the closest record to their discharged date}, in the processed dataset.
    \item On the contrary to the author's claim, this approach is operationally infeasible. During training, retrospective models are built upon \textit{patients’ last records} in the hospital, but in the real world, we will never know in advance if a record is one’s \textit{last record}. Therefore, the model is inapplicable due to a potential data distribution difference between training samples and real-world testing samples.
    \item It is reasonable for the model built and tested using only \textit{the most recent data record which is the closest to the discharged date}, or, the last records, to have better performance than models built and tested otherwise, because of the continuity of time, samples tend to be more separable as they approach closer to their discharged date/death date
\end{itemize}
These discussions around potential improvements on the original work inspire our definition of the deterioration problem as well as the proposed method to solve it. Instead of the retrospective version of problem definition stated in Problem Statement 0, we propose the predictive version of the COVID-19 deterioration problem.

\textbf{Problem Statement 1}\label{stmt:one} (Predictive Modeling of COVID-19 Deterioration):Assuming there are $n_{p}$ hospitalized COVID-19 patients. For patient $i$, $1 \leq i \leq n_{p}$, the daily feature vector is $\mathbf{ x_{t,i}}$. Admission and discharged datetimes are defined as $T_{a,i}$ and $T_{e,i}$. The goal is to build binary classifier $C$ which minimizes the loss function:
\[ \mathcal{L} = \Omega(C) + \sum_{i=1}^{n_{p}} \sum_{t=T_{a,i}}^{T_{e,i}} ce(C(\mathbf{ x_{t,i}}), y_{i})\] 
where $\Omega(C)$ denotes the regularization loss and $ce$ is the cross-entropy.

As we can see, instead of building models upon \textit{last records}, the new problem statement will work for every patient every day. What’s more, for patients whose conditions are progressively worsening, in order to reflect the increasing importance of near-term predictions, we can assign weights $\lambda_{t}$ to days in the loss function, i.e.,$\mathcal{L} = \Omega(C) + \sum_{i=1}^{n_{p}}  \sum_{t=T_{a,i}}^{T_{e,i}} \lambda_{t} ce(C(\mathbf{ x_{t,i}}), y_{i})$ ,so that it will regularize the classifier toward more accurate near-term predictions.

\section{\textbf{Time-varying effects of clinical variables}}\label{sec3}
We motivate our methods by showing the evidence for the following assumption:\\
\textbf{Assumption 1}: Clinical variables have time-varying effects on COVID-19 deterioration outcomes.

Some features may be important for early stage COVID-19 patients and others may become important later as the disease progresses. To investigate the validity of the assumption, we will conduct the following experiments. 

\begin{figure}[h]
  \centering
  \includegraphics[width=\linewidth]{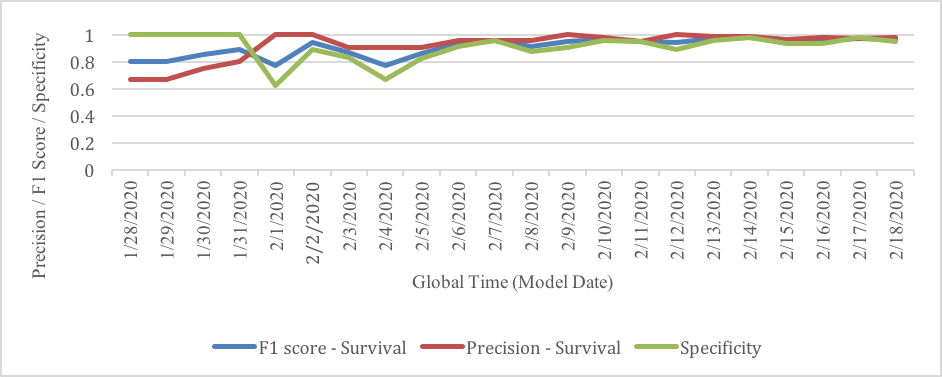}
  \caption{Baseline daily models using global time – Accuracy for Patient Survival }
  \label{fig:2a}
\end{figure}

\begin{figure}[h]
  \centering
  \includegraphics[width=\linewidth]{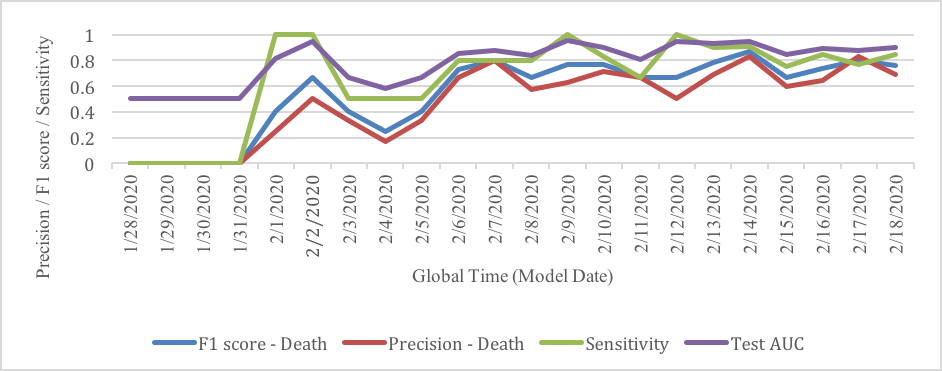}
  \caption{Baseline daily models using global time – Accuracy for Patient Death}
  \label{fig:2b}
\end{figure}

\cite{yan2020interpretable} found three key features, namely, lactic dehydrogenase (LDH), lymphocytes and high-sensitivity C-reactive protein (hs-CRP), as most important features in the model. Using the \textit{Wuhan Dataset}, we built models for each date, where the input to the model was the latest laboratory result values of the three key features for patients currently admitted to the hospital. A supervised XGBoost classifier\cite{chen2016xgboost} was used similar to the one used in \cite{yan2020interpretable}. Each model was trained on the training set data, and then we tested the performance of the model in predicting patient mortality on an external test set. Fig. \ref{fig:2a} and \ref{fig:2b} show the performance of the models for predicting survival and death respectively. As can be seen from these figures, the baseline models using latest available laboratory results were able to predict patient survival with high accuracy within few days of being admitted to the hospital. But the models were random at best in predicting mortality. 

\begin{figure}[h]
  \centering
  \includegraphics[width=\linewidth]{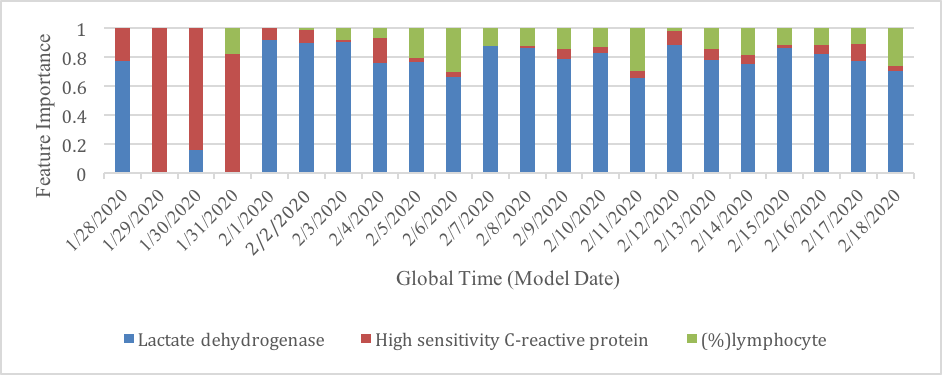}
  \caption{Feature Importance for each baseline model}
  \label{fig:2c}
\end{figure}

Moreover, Fig.\ref{fig:2c}, which shows feature importance for each of these models, strengthens our assumption that the importance of features may change with disease progression. Although further investigations with more data and hence, more statistical power, are needed to substantiate our findings, nevertheless, the analyses above clearly demonstrate the value and possibility to improve from the ``one model for all'' approach.

\section{\textbf{Temporal Stratification }}\label{sec4}
\subsection{\textbf{Patient Alignment}}
When patients are admitted to the hospital, they are at different stages in the course of COVID-19. What's more, due to demographics, comorbidities and other risk factors, the rate of disease progression or recovery may vary from person to person. In a sense, this implies that the clinical data from these patients is heterogeneous. Moreover, as Assumption 1 states, clinical variables may play time-varying roles during the entire course of a disease. Thus, learning one model for all progression stages of a disease may not be the best solution. To tackle this issue, we may learn different models for each stage. First, patients' time series data is aligned by a well-defined reference time point, such as days of symptom onset or days to outcome. Next, patient data can be either temporally binned, or stratified according to clinical definitions. The goal is to roughly segment one patient stay into separate stages. Finally, given enough data, predictive models are developed for each stratum.

\begin{figure}[h]
  \centering
  \includegraphics[width=\linewidth]{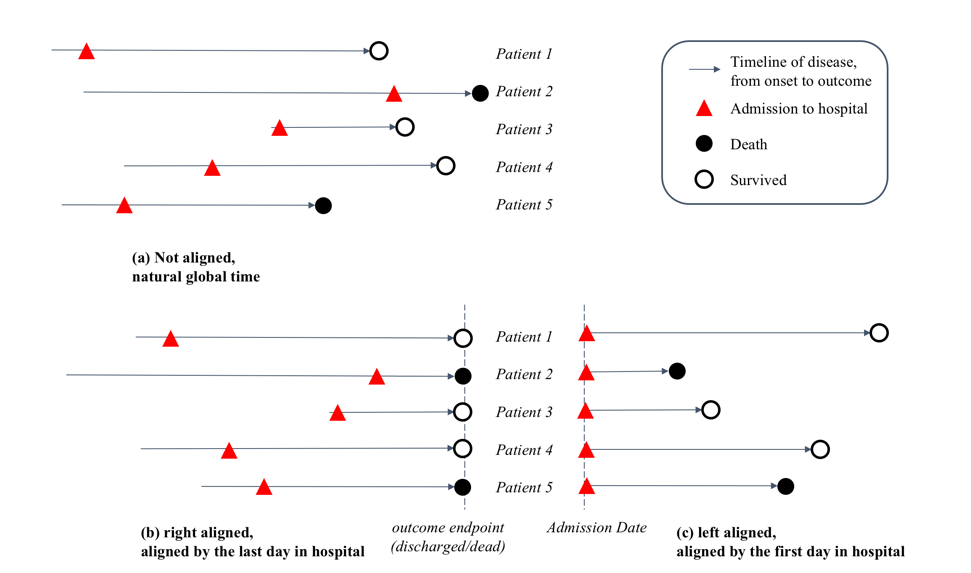}
  \caption{Alignment methods}
  \label{fig:3}
\end{figure}

Fig.\ref{fig:3} lists two common patient alignment methods, (b) aligned by the endpoint/discharged date, (c) aligned by the first day of admission, as well as (a) the natural calendar time. Though aligning by symptoms onset or day of contact would be an ideal way, especially for an epidemiology study; for infectious diseases like COVID-19, during a global massive pandemic, it is not feasible to accurately trace and track the symptom onset date or exact contact date for every patient. Therefore, it is reasonable to choose patients’ discharged/outcome date as a reference time point. In this paper, patients are aligned by their outcome days and subsequently temporally stratified.

\subsection{\textbf{Stratification-Prediction Framework}}
In light of previous sections, the operationally feasible solution to Problem Statement 1 should also address Assumption 1 and preferably, be able to be incorporated into other important prediction goals such as length of stay. While the method in this section is developed for prediction of COVID-19 patients’ clinical deterioration while hospitalized, the general framework should also apply to other diseases, as long as the assumption still holds. The proposed method should work with both general wards and ICU patients, even though ICU data is expected to be denser and more varied. For simplicity and consistency with the original \textit{Wuhan Dataset}, we won't distinguish between general ward patients and ICU patients. We begin with discussions around the patient alignment problem, then move on to the proposed method, Temporally Stratified Model.

\begin{figure}[H]
  \centering
  \includegraphics[width=\linewidth]{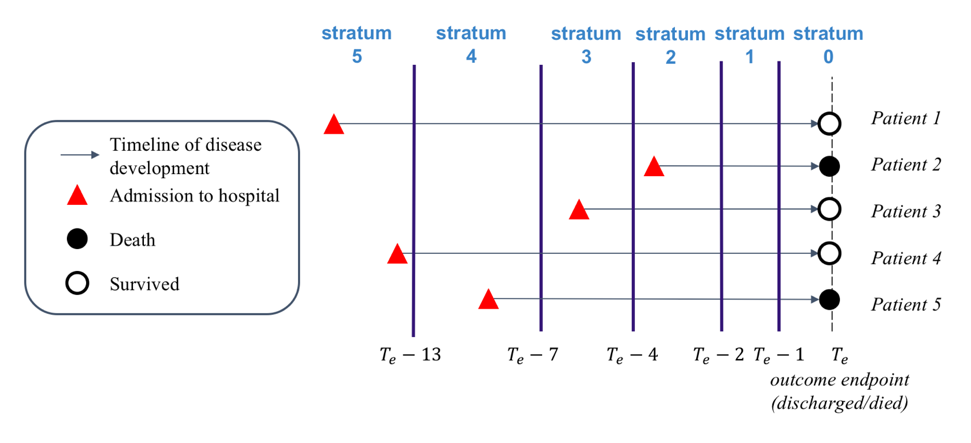}
  \caption{Temporal stratification }
  \label{fig:4}
\end{figure}

Fig.\ref{fig:4} shows that after aligning patients by discharged/outcome dates, we binned their time series data into predefined strata. The strata can be thought of as hidden states that control the distribution of clinical variables during a period of time. As a general framework, we can arbitrarily choose clinically meaningful strata definitions as long as the label would fit into a multi-class classification problem. In this pilot study, we stratified patients by their remaining length of stay. Therefore, the task of stratification is now effectively equivalent to the task of predicting the remaining length of stay.  By right-aligning COVID-19 patients and temporally stratifying time series data, we decomposed the Problem Statement 1 into the following two sub-problems, Strata Assignment and Stratum-wise deterioration Prediction. Though there are algorithms which can implicitly model latent states, we chose to explicitly model the inferring of strata/hidden states as Strata Assignment.

As Fig.\ref{fig:4} demonstrates, strata are mutually exclusive periods of time, yet collectively cover all days. Formally, let $\mathcal{S}$ denote a set of time points delimiting neighboring stratum, then $ \mathcal{S} = {d_{1},d_{2}, \dots , d_{n_{s}-1}}$, where $0 \geq d_{1} \geq d_{2} \geq \dots \geq d_{n_{s}-1}$. $|\mathcal{S}| = n_{s}-1$ since $n_{s}-1$ points are needed to partition the timeline into $n_{s}$ segments. In Fig.\ref{fig:3}, $\mathcal{S}={-1,-2,-4,-7,-13}$.  To be more specific, given $\mathcal{S}$, strata are defined as the following half-open half-closed time periods, $s_{1} = (-\infty, T_{e}+d_{n_{s}-1}], s_{n_{s}-2} = (T_{e}+d_{2}, T_{e} + d_{1}], \dots, s_{n_{s}-1} = (T_{e}+d1, T_{e})$
Notice that the data on the last day are discarded during training and testing since it would be too late for physicians to take any actions on the patients. Since all data are right-aligned, we can omit $T_{e}$ and then the notations would become $s_{1} = (-\infty, d_{n_{s}-1}], s_{n_{s}-2} = (d_{2}, d_{1}], \dots, s_{n_{s}-1} = (d1, 0)$

\textbf{Problem Statement 2}\label{stmt:two} (Patient Strata Assignment) Assuming strata $\mathcal{S}$, given daily clinical variables $\mathbf{ x_{t}}$, learn a multi-class stratification classifier $\mathcal{F}$ which determines a point mass probability function over $n_{s}$ strata, namely 
$\mathcal{F}(\mathbf{ x_{t}}) = \mathbf{ p_{t}}= [p_{t}^{1}, p_{t}^{2}, \dots , p_{t}^{n_{s}} ]$, the predicted probability of being in each stratum. We have $\sum_{k=1}^{n_{s}}p_{t}^{k} = 1$, $p_{t}^{k} \in [0,1]$, for $0 \leq k \leq n_{s}$.Since, we designate the Strata Assignment task to be a daily task for every patient, the loss function can be written as: 

\[ \mathcal{L} = \Omega(\mathcal{F}) + \sum_{i=1}^{n_{p}} \sum_{t=T_{a,i}}^{T_{e,i}} ce(\mathcal{F}(\mathbf{ x_{t,i}}), s_{t,i})\]

\textbf{Problem Statement 3}\label{stmt:three} (Stratum-wise Deterioration Prediction) Assume there is a solution $\mathcal{F}$ to the Strata Assignment problem. For every stratum $k = [a,b) \in \mathcal{S} $, for any patients who have data during this stratum defined period of time, learn a binary classifier $M^{k}$ that takes feature vectors and outputs a probability score $\hat{y}_{t}^{k} = Pr(\hat{y}_{t}|\hat{s}_{t}=k) = M^{k}(\mathbf{ x_{t}})$, for $t \in [a,b)$. The loss function for stratum k can be written as:
\[ \mathcal{L}^{k} = \Omega(M_{k}) + \sum_{patients \in s_{k}} \sum_{t=max(T_{a},b-1)}^{a} ce(M_{k}(\mathbf{ x_{t,i}}), y_{i})  \]

If there are $n_{s}$ strata, there will be $n_{s}$ stratum-wise deterioration prediction models. We now describe all the necessary steps of the proposed solution to Problem 1, 2 and 3. As illustrated by Fig.\ref{fig:5a}, given any strata definition $\mathcal{S}$, proceed with the following steps:

\begin{figure}[h]
  \centering
  \includegraphics[width=\linewidth]{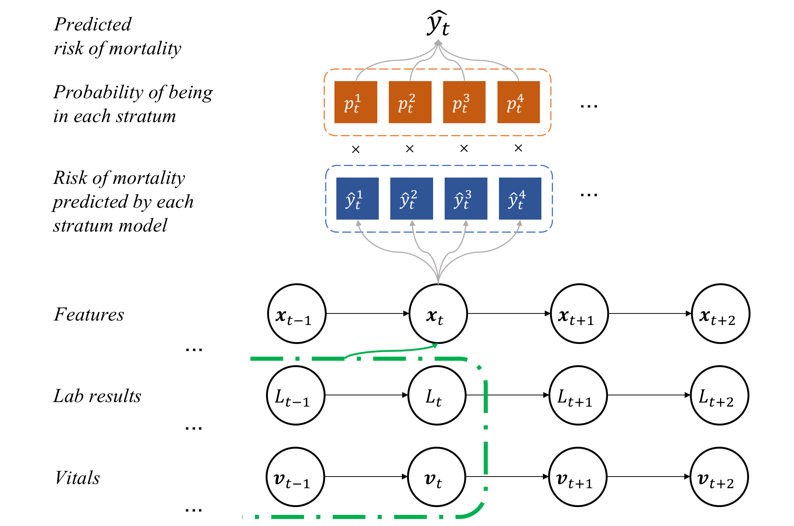}
  \caption{Calculate risk scores}
  \label{fig:5a}
\end{figure}

\begin{enumerate}[label=\alph*.]
\item Temporally stratify patients into $n_{s}$ strata. The strata definition can be either given by medical experts or derived from the data.
\item Build strata classifier $\mathcal{F}$ to assign stratum for every COVID-19 patient every day. If temporally stratified, it is equivalent to predict the remaining length of stay.
\item Build stratum-wise classifiers for every stratum $(a,b]$. A double truncation is implied here, meaning the training set for a particular stratum only contains those patients who survived at least $a$ days (left truncation) and had valid data during this stratum period(right truncation).
\item When testing, the outcome is given by the weighted average of each stratum-wise model, namely $\hat{y}_{t} = \mathbf{p_{t}} \cdot M_{t} $
\end{enumerate}

The risk of mortality prediction $\hat{y}_{t}$ is produced by the dot product of strata probability vector $\mathbf{p_{t}}$ and the vector of predicted outcomes from $n_{s}$ stratum-wise models, $M_{t} = [M^{1}(\mathbf{ x_{t}}), M^{2}(\mathbf{ x_{t}}), \dots , M^{n_{s}}(\mathbf{ x_{t}}) ]$ ,where $M^{k}(\mathbf{ x_{t}})$ is the prediction made by the $k-th$ stratum risk prediction model. One issue with this approach is the choice and definitions of temporal strata, which affects the number of strata needed. There is no definite answer to these questions. To determine a proper definition, we can rely on medical knowledge, any reasonable heuristics, or even unsupervised time series clustering algorithms. As for the problem of determine the number of strata, we propose that given there is enough data within each stratum, having more strata than necessary will not significantly decrease the performance in terms of the total loss $\mathcal{L}$. In the worst case scenario, stratum-wise deterioration classifiers are similar to each other and is equivalent to the unstratified model trained using all data.

\section{\textbf{Experiments}} \label{sec5}

\subsection{\textbf{Wuhan Dataset}}
The \textit{Wuhan Dataset} belongs to COVID-19 patients from the region of Wuhan, China and is from a mixed population of hospitalized patients in general wards (mild, severe) and ICU (critical). There are 375 patients in the training set, where 201 patients eventually recovered, while 174 died (46\%). For each patient, input consists of values for 76 variables (lab results). Table \ref{tab:wuhanTrain} and Table\ref{tab:wuhanTest} provide more details. An additional 110 patients (12\% died) are in the test set, but that dataset has information for only 3 laboratory results.

\begin{table*}
  \caption{\textit{Wuhan Dataset}, Training set}
  \label{tab:wuhanTrain}
  \begin{tabular}{lrrrrl}
    \toprule
    & \textbf{Critical}   & \textbf{Severe}   & \textbf{Mild}   &  \textbf{Total} & \\
    \midrule
   \textbf{Recovered}  & 3 (0.8\%) & 13  (3.47\%) & 185  (49.33\%) & 201  (53.60\%)  \\
    \textbf{Died} & 148  (39.47\%) & 14  (3.73\%) & 12   (3.2\%)   & 174  (46.4\%)   \\
    \textbf{Total} & 151 (40.27\%) & 27  (7.20\%) & 197  (52.53\%) & 375  (100.00\%) \\
    \bottomrule
  \end{tabular}
\end{table*}

\begin{table*}
  \caption{\textit{Wuhan Dataset}, Test set}
  \label{tab:wuhanTest}
  \begin{tabular}{lrrrrl}
    \toprule
     & \textbf{Critical}   & \textbf{Severe}   & \textbf{Mild}   &  \textbf{Total} & \\
    \midrule
    \textbf{Recovered}  & 13  (11.82\%) & 34 (30.91\%)  & 50  (45.45\%)  & 97  (88.18\%)   \\
    \textbf{Died}       & 9   (8.18\%)  & 4  (3.64\%)   & 0   (0.00\%)   & 13  (11.82\%)   \\
    \textbf{Total}      & 22  (20\%)    & 38 (34.55\%)  & 50  (45.45\%)  & 110 (100.00\%)  \\
    \bottomrule
  \end{tabular}
\end{table*}

\begin{figure}[h]
  \centering
  \includegraphics[width=\linewidth]{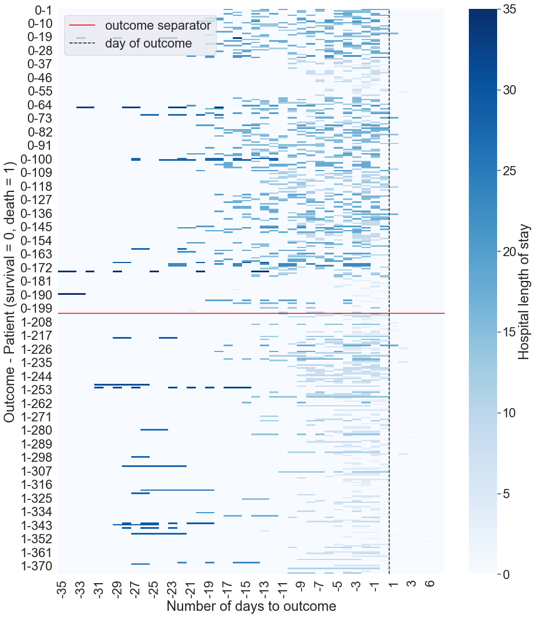}
  \caption{Right-aligned \textit{Wuhan Dataset} to the outcome for patients (discharge from the hospital or death). The red horizontal line is the outcome separator. Rows above the red line show data availability for recovered patients. Rows below the red line show data availability for deaths. The darker color represents a longer total length of stay.}
  \label{fig:wuhandataset}
\end{figure}

Fig.\ref{fig:wuhandataset} shows the right-aligned data in the \textit{Wuhan Dataset}. Each patient had medical tests performed at different points in time during their hospital stay, and every day when such data was available is shown as a darker point based on time to outcome. All stays have been right aligned. Moreover, the length of stay is indicated by the darkness of color marked. For example, patient 252 stayed in the hospital for 32 days and had data available at days -31, -29, -27, -23 etc. The outcome for this patient was death. As another example, patient 54 stayed in the hospital for 7 days, had data available on -6, -3 and 0 days, and had an outcome of survival. Another observation from this figure is that patients who did not survive had shorter stays in the hospital when compared to those who survived, which can be seen from the darker areas in the top half of the figure. Notice that the density of tests is much closer to the right side, which illustrates the need to have a more accurate estimate of the patient’s condition closer to the end of stay. This intuition is the basis for having time strata of differing lengths as shown in Fig.\ref{fig:4}.

\subsection{\textbf{Results and Analysis}}
We built supervised classifiers using an XGBoost algorithm\cite{chen2016xgboost}, which is a tree-based gradient boosting algorithm. Models were built for each stratum, $\mathcal{S} = {-1,-2,-4,-7,-13}$, which takes as input, all data that was associated with days within the stratum. For example, the model for stratum -1 would get trained on patient data in days range  $(-2 ,-1]$. Similarly, the model for stratum -13 would use all data from date of admission up to 13 days before the outcome, i.e. $(-\infty ,-13]$ for training. Prediction by model 
$M^{k}(\mathbf{ x_{t}})$ in the $k-th$ stratum is a probability of death within $k$ days. Various configurations of XGBoost were used to train the models, and the configurations with the best outcomes on the validation set were used to make predictions on the test set. Here, we predefined the strata definition to strike a balance between having enough number of available patients in each strata and modeling the underlying disease progression stages pointed by medical literature\cite{garg2020hospitalization, bellino2020covid, wu2020risk, polak2020systematic}.

\begin{figure}[h]
  \centering
  \includegraphics[width=\linewidth]{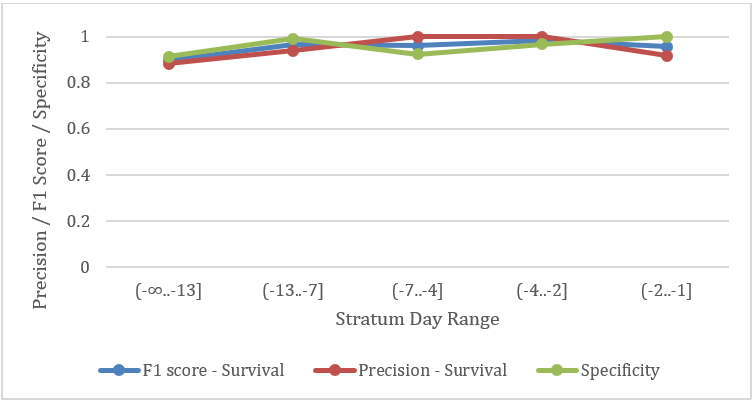}
  \caption{Stratum-wise Models – Accuracy for Patient Survival}
  \label{fig:6a}
\end{figure}

\begin{figure}[h]
  \centering
  \includegraphics[width=\linewidth]{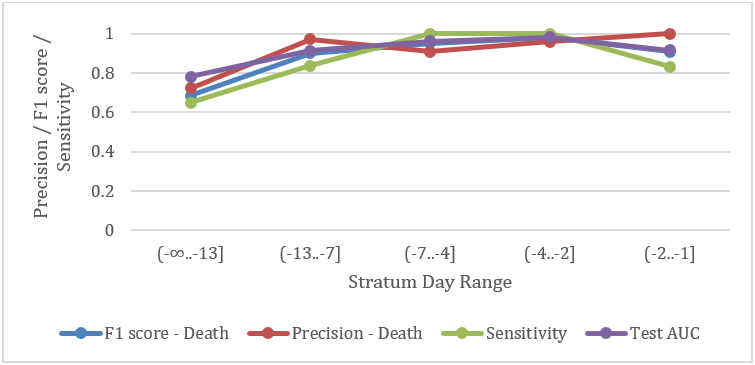}
  \caption{Stratum-wise Models – Accuracy for Patient Death}
  \label{fig:6b}
\end{figure}

\begin{figure}[h]
  \centering
  \includegraphics[width=\linewidth]{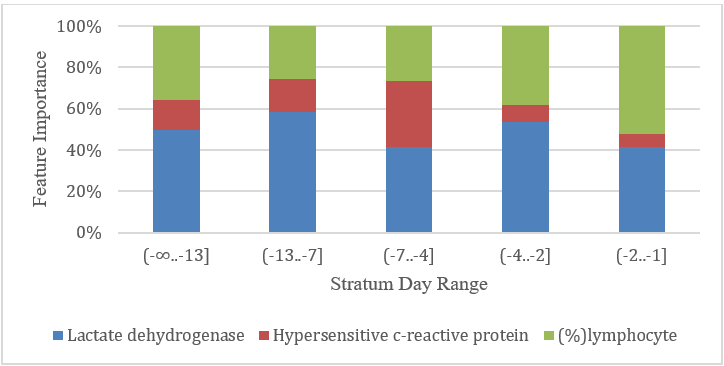}
  \caption{Feature Importance for Stratum-wise Models}
  \label{fig:6c}
\end{figure}

Compared to the baseline models in Fig.\ref{fig:2a}, Fig.\ref{fig:2b}, and Fig.\ref{fig:2c}, the stratum-wise models(results shown in Fig.\ref{fig:6a}, Fig.\ref{fig:6b}, Fig.\ref{fig:6c}, and Table\ref{tab:strata-model}) , perform consistently for death outcome, with precision close to 1 about 2 weeks before the outcome and sensitivity close to 1 about 1 week before the outcome. All experiments are coded in python. We use XGBoost for both stratum classifiers and stratum-wise prediction models. The stratum-wise model use decision trees as base classifiers. Hyper-parameters are tuned based on a 20\% split(validation set) within the training set. The maximum depth of each base classifier is set to be 4. Both stratum classifiers and stratum-wise prediction models are trained with a learning rate of 0.2 and a subsample rate of 0.9. No L1 or L2 regularization for stratum-wise classifiers. We set a 0.02 L2 regularization for strata classifier. 

\begin{table*}
  \caption{Performance of the Stratum-wise Models}
  \label{tab:strata-model}
  \begin{tabular}{rrr|ccc|cccc}
    \toprule
     & Statistics & & $y=1$ & means survived &  & $y=1$ & means died &  & \\
     \textbf{Stratum} & \textbf{\#Patients} & \textbf{\# Deaths} & \textbf{F1 score} & \textbf{Precision} & \textbf{Specificity} & \textbf{AUC ROC} & \textbf{F1 score} & \textbf{Precision} & \textbf{Sensitivity}\\
    \midrule
    $(-\infty, -13]$ & 46 & 7 & 0.897 & 0.881 & 0.912 & 0.781 & 0.684 & 0.722 & 0.650\\
    $(-13, -7]$ & 99 & 13 & 0.963 & 0.938 & 0.991 & 0.914 & 0.900 & 0.973 & 0.837\\
    $(-7, -4]$ & 48 & 12 & 0.960 & 1.000 & 0.923 & 0.962 & 0.952 & 0.909 & 1.000\\
    $(-4, -2]$ & 40 & 12 & 0.983 & 1.000 & 0.967 & 0.983 & 0.979 & 0.958 & 1.000\\
    $(-2, 1)$ & 34 & 12 & 0.957 & 0.917 & 1.000 & 0.917 & 0.909 & 1.000 & 0.833\\
    \bottomrule
  \end{tabular}
\end{table*}

To evaluate the performance of the combined models (strata assignment and stratum-wise prediction), we split the dataset twice to generate training cohort (60\% of the patients), validation cohort (20\% of the patients) and the test cohort (20\% of the patients). In order to estimate the standard error, the stratified group 5-fold cross validation is repeated 100 times.  The AUROC is 0.9805 with a standard error of 0.0198. More details are shown in Table\ref{tab:combined-model}. 

We also tried to use the mean, slope, standard deviations and the first order difference of lab results. However, due to the sparseness of the data, additional features show little or no improvements. For a particular lab test, during the entire stay, one patient may only take 2-3 times. To some extent, the sparseness of lab results prevents the usage of more complicated sequential models like hidden Markov models\cite{rabiner1986introduction} and recurrent neural networks\cite{hochreiter1997long}.

\begin{table*}
  \caption{Performance of the Combined Models}
  \label{tab:combined-model}
  \begin{tabular}{llllll}
    \toprule
    \textbf{AUC ROC}(se) & \textbf{AUC PR}(se) & \textbf{Accuracy}(se) & \textbf{F1 score}(se) & \textbf{Precision}(se) & \textbf{Recall}(se) \\
    \midrule
    0.9805 & 0.9720 & 0.9353 & 0.9134 & 0.9186 & 0.9105\\
    (0.0198) & (0.0142) & (0.0225) & (0.0344) & (0.0469) & (0.0479)\\
    
    \bottomrule
  \end{tabular}
\end{table*}

\section{\textbf{Discussions}} \label{sec6}

One benefit of the temporal stratification approach is that it incorporates the change of feature importance across different stages. Another advantage is that it allows a multi-task learning of outcome predictions and resources usage predictions, since the stratum contains information on how long patients will continue to stay. In the temporal stratification approach, the remaining length of stay is implicitly modeled and predicted by the stratum classifier. Problem 2 is equivalent to predict the remaining length of stay. Still, the definition of strata remains somehow arbitrary and is yet to be explored. Besides temporal stratification, we can also use predefined medical phenotypes, or unsupervised clustering algorithms to identify latent states. Furthermore, we can impose restrictions on transitions between strata based on medical professional’s advice. The idea is to learn the state-space model which governs the progression of diseases. 

There are also several limitations of this study. First of all, the method for deriving strata definition is somehow heuristic. One can have more or less strata, based on the size of the dataset and the nature of  the disease. An ideal strata definition should satisfy two criterion: (i)each stratum should have enough data and patients, so that k-fold cross validation can be conducted during model training. (ii) Breakdown points should be carefully determined in order to approximate the actual disease progression trajectory.
Secondly, there are limitations of the dataset, namely the test set does not have as many variables as the training set does. Therefore, we are not able to utilize the full 100 variables on both training and testing sets. What’s more, the dataset only contains sparse lab variables, while medical history, demographics and vitals are not published. For a particular lab test, during the entire stay, one patient may only take 2-3 times. To some extent, the sparseness of lab results prevents the usage of more complicated time-dependent models. Last but not least, variations in patients and disease severity groups are not taken into account. Comorbidities, ages, and gender are often confounders which affects the final outcome and also importantly, the speed of disease progression or recovery. The stratification-prediction approach is assuming all the patients who finally died of COVID are approximately following the same progression trajectory, of which the variation is controlled by the width of each stratum. However, this may not hold in the real case or in other diseases.

For future work, we consider using sequential models and applications to other diseases and prediction targets, such as ICU admissions and the use of ventilators. We list possible extensions here.
\begin{itemize}
    \item \textbf{Multitask learning.} Problem 2 and Problem 3 can be either solved by different machine learning algorithms, or trained jointly using any multitask learning models. If they are modeled jointly, the loss function is given by $\mathcal{L} = \mathcal{L}_{s} + \sum_{k=1}^{n_{s}}\mathcal{L}^{k}$. Further if problem 2 is modeled as an auxiliary task, it is possible to tweak the loss function by adding a coefficient to $\mathcal{L}_{s}$. Otherwise the loss function suggests a multitask learning implementation, such as deep neural networks.
    
    \item \textbf{Flexible strata definition.} Besides temporal stratification, we can use predefined medical phenotypes, or clustering algorithms to identify latent states. What’s more, we can impose restrictions on transitions between strata based on medical professional’s advice. 
    
    \item \textbf{Incorporating existing triage models.} Noting that, if instead of daily reassignment, the stratification is only done once at admission, this is equivalent to the triaging/risk stratification problem. There is little data available at one’s initial time period during a hospital stay, the problem is known as a cold start problem. In this case, we can utilize existing triage protocols or treat this as a separate problem.

    \item \textbf{Markov vs. Non-Markov Approach.} Current implementation only uses the latest data while ignoring the EHR history. $\mathbf{ x_{t}}$ denotes the feature vector, as shown in Fig.\ref{fig:5a}, it can certainly include the features calculated from history readings of lab results and vital signs. 
    
\end{itemize}

\begin{figure*}[h]
  \centering
  \includegraphics[width=15cm]{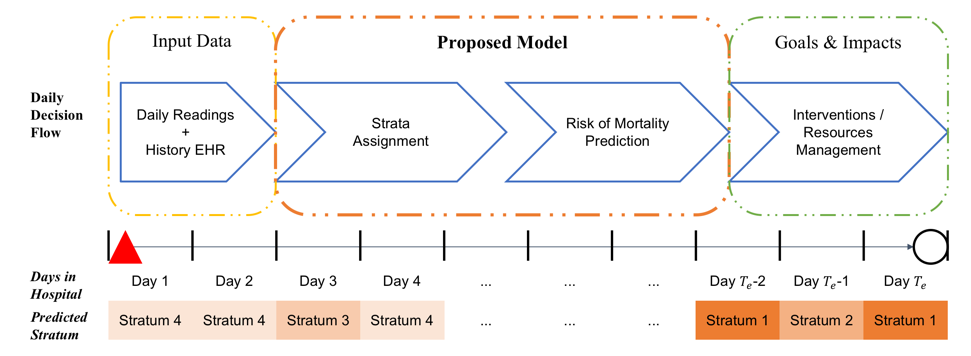}
  \caption{Proposed workflow for temporally stratified models}
  \label{fig:5b}
\end{figure*}

The proposed workflow for the deterioration prediction problem is demonstrated in Fig.\ref{fig:5b}. The method is a flexible and general method which can be applied to not only COVID-19 but also many other diseases. For ICU patients, instead of making daily predictions, we can adjust the prediction frequency to hourly. For other diseases, stratum can be defined as some phenotypes or latent states of the disease. 

\newpage

\section{\textbf{Conclusions}}
In this pilot study, we verify the assumption that clinical variables are likely to have time-varying effects on COVID-19 deterioration outcomes. To address this issue, we propose temporally stratified models, and effectively delineate the deterioration prediction problem into two sub-problems: strata assignment and stratum-wise risk of mortality prediction. The method can be generalized as a stratification-prediction approach framework. Prototype models are developed using the \textit{Wuhan Dataset}. Our experiments show promising results, encouraging further investigations. We are currently collecting data for COVID-19 hospitalized patients admitted into a large hospital system in an IRB–approved study. In the future, we plan to apply the proposed method to other disease and prediction targets, derive models using this hospital based dataset, and validate models in a future prospective study.

\begin{acks}
We would like to thank the authors of \textit{Wuhan Dataset}\cite{yan2020interpretable}  for their decision to make the dataset publicly available. 
\end{acks}

\bibliographystyle{ACM-Reference-Format}
\bibliography{references}

\end{document}